\documentclass[sigconf]{aamas}  

\usepackage{subfig}
\usepackage{url}
\usepackage{microtype}
\usepackage{tipa}
\usepackage{mathtools}
\usepackage{float}
\restylefloat{figure}
\usepackage{amsmath}
\usepackage{amssymb}  
\usepackage{mathtools}
\usepackage{fancyhdr}
\usepackage{kantlipsum}
\usepackage{eso-pic}
\usepackage{nomencl}
\usepackage{etoolbox}
\usepackage{balance}
\usepackage{dsfont}
\usepackage{tikz}
\usepackage{eucal}
\tikzset{>=latex}
\usetikzlibrary{automata,positioning}
\usepackage[algo2e,linesnumbered,ruled,lined]{algorithm2e}
\usepackage{booktabs}

\newcommand\x{0.21}
\newcommand{\edit}{\textcolor{black}}

\setcopyright{rightsretained}  
\acmDOI{doi}  
\acmISBN{}  
\acmConference[]{}{}{}  
\acmYear{2019}  
\copyrightyear{2019}  
\acmPrice{}  

\begin{document}

\title{Logically-Constrained Neural Fitted Q-iteration}  



\author{$~$\\$~$\\Mohammadhosein Hasanbeig, Alessandro Abate, and Daniel Kroening}  

%
\affiliation{%
	\institution{Computer Science Department, University of Oxford}
	\streetaddress{Parks Road}
	\city{Oxford} 
	\state{United Kingdom} 
	\postcode{OX1 3QD}
}
\email{[hosein.hasanbeig, alessandro.abate, daniel.kroening]@cs.ox.ac.uk}
\email{}
\email{}
\email{}
%
%
%
%
%
%
%

\begin{abstract}  
We propose a method for efficient training of Q-functions for
continuous-state Markov Decision Processes (MDPs) such that the traces of
the resulting policies satisfy a given Linear Temporal Logic (LTL) property. 
LTL, a modal logic, can express a wide range of time-dependent logical
properties (including \emph{safety}) that are quite similar to patterns in natural language. 
We~convert the LTL property into a limit deterministic B\"uchi automaton and
construct an on-the-fly synchronised product MDP.  The control policy is
then synthesised by defining an adaptive reward function and by applying a modified neural fitted Q-iteration algorithm
to the synchronised structure, assuming that no prior knowledge is available
from the original MDP.  The proposed method is evaluated in a numerical
study to test the quality of the generated control policy and is compared
with conventional methods for policy synthesis such as MDP abstraction
(Voronoi quantizer) and approximate dynamic programming (fitted value
iteration).
\end{abstract}



\maketitle

\section{Introduction}
\label{sec:intro}

Reinforcement Learning (RL) is a promising paradigm for training an
autonomous agent to make optimal decisions when interacting with an MDP if
the stochastic behaviour of the MDP is initially unknown. However,
conventional RL is mostly focused on problems in which the set of states of
the MDP and the set of possible actions are finite. Nonetheless, many
interesting real-world problems require actions to be taken in response to
high-dimensional or real-valued sensory inputs~\cite{doya}.  As an exemplar,
consider the problem of drone control, in which the drone state is
represented as its Euclidean position $(x,y,z) \in \mathds{R}^3$: the
state space of an MDP modelling the stochastic behaviour of the drone is
uncountably infinite, namely continuous.

The simplest way to apply RL to an infinite-state MDP is to discretise the
state space of the MDP a-priori and to fall back to conventional RL with a
state-action-reward look-up table in order to find the optimal
policy~\cite{cdp}. Although this method works well for many problems, the
resulting discrete MDP is often inaccurate and may not capture the full
dynamics of the original MDP. Thus, discretisation of MDPs suffers from the
trade off between accuracy and the curse of dimensionality.

A smarter solution is to collect a number of samples and only then apply an
approximation function that is constructed via regression over the set of
samples. The approximation function replaces the conventional RL
state-action-reward look-up table by generalising over the state space of
the MDP. A number of approximation methods are available, e.g.,
CMACs~\cite{cmac}, kernel-based modelling~\cite{ormo}, tree-based
regression~\cite{ernst}, basis functions~\cite{babuska}, etc.  Among these
methods, neural networks stand out because of their ability to approximate
any non-linear function~\cite{apprx}.  Numerous successful applications of
neural networks in RL for infinite or large-state space MDPs have been
reported, e.g., Deep Q-networks~\cite{deepql}, TD-Gammon~\cite{tesauro},
Asynchronous Deep RL \cite{asynchronous}, Neural Fitted
Q-iteration~\cite{nfq} and CACLA~\cite{cacla}.

In this paper, we propose to employ multi-layer perceptrons to synthesise a
control policy for infinite-state MDPs such that the generated traces
satisfy a Linear Temporal Logic (LTL) property.  LTL allows the
formalization of complex mission requirements in a rich time-dependent
language.  By employing LTL we are able to express sophisticated high-level
control objectives that are hard to express and achieve for other methods
from vanilla RL~\cite{sutton,smith} to more recent developments such as
Policy Sketching~\cite{pol-sketch}.  Examples include liveness and cyclic
properties, where the agent is required to make progress while concurrently
executing components to take turns in critical sections or to execute a
sequence of tasks periodically.  \edit{We show that the proposed
architecture is efficient and is compatible with RL algorithms that are core
of recent developments in the community.}

\edit{To the best of our knowledge, no prior research has been done that
enables RL to generate policies that satisfies an arbitrary given LTL
property for \emph{continuous-state} MDPs.  
So far, Logic-based synthesis over such models has been limited to DP- or optimisation-based techniques \cite{AKLP10,HSA17,tkachev}. 
By contrast, the problem of
control synthesis for \emph{finite-state} MDPs for temporal logic has
received a lot of attention.  In~\cite{wolf}, the property of interest is
given in LTL and is converted into a Deterministic Rabin Automaton (DRA). 
A~modified Dynamic Programming (DP) algorithm then maximises the worst-case
probability of satisfying the specification over all transition
probabilities.  Note that in this work the MDP must be known a priori. 
\cite{topku} and \cite{brazdil} assume that the given MDP has unknown
transition probabilities and build a Probably Approximately Correct MDP (PAC
MDP), which is synchronised via production with the logical property after conversion to DRA. 
The goal is to calculate the value function for each state such that the
value is within an error bound of the actual state value where the value is
the probability of satisfying the given LTL property.  The PAC MDP is
generated via an RL-like algorithm and standard value iteration is applied
to calculate the values of states.} The first model-free RL algorithm that
is able to synthesise policies for a fully unknown MDP such that an LTL
formula is satisfied appears in~\cite{arxiv}.  Additionally, \cite{arxiv}
shows that the RL procedure sets up a local value iteration method to
efficiently calculate the maximum probability of satisfying the given
property, at any given state of the MDP.  An in-depth discussion
of~\cite{arxiv} with extended proofs and experiments can be found
in~\cite{journal_arxiv}.  The work in~\cite{arxiv} has been taken up more recently by~\cite{hahn}, 
which has focused on model-free aspects of the algorithm and has employed a different LDBA structure and reward.   

\edit{Moving away from full LTL logic, scLTL is proposed for mission
specifications, with which a linear programming solver is used to find
optimal policies.  The concept of shielding is employed in~\cite{shield} to
synthesise a reactive system that ensures that the agent stays safe during
and after learning.  However, unlike our focus on full LTL expressivity,
\cite{shield} adopted the safety fragment of LTL as the specification
language.  This approach is closely related to teacher-guided
RL~\cite{teacher}, since a shield can be considered as a teacher, which
provides safe actions only if absolutely necessary.  The generated policy
always needs the shield to be online, as the shield maps every unsafe action
to a safe action.  \cite{fulton} and \cite{fulton2} address safety-critical
settings in RL in which the agent has to deal with heterogeneous set of
environments in the context of cyber-physical systems.  \cite{fulton3}
further employs DDL \cite{ddl}, a first-order multimodal logic for
specifying and proving properties of hybrid programs.  Almost all other
approaches in safe RL either rely on ergodicity of the underlying MDP,
e.g.~\cite{abbeel}, which guarantees that any state is reachable from any
other state, or they rely on initial or partial knowledge about the MDP,
e.g.~\cite{initial1} and~\cite{initial2}.}

\section{Background}
\label{background}
\subsection{Problem Framework} 
\begin{definition}[Continuous-state Space MDP]\label{mdpdef} The tuple $\textbf{M}=(\allowbreak
	\mathcal{S},\allowbreak\mathcal{A},\allowbreak s_0,\allowbreak
	P,\allowbreak\mathcal{AP},\allowbreak L)$ is an MDP over a set of
	states $\mathcal{S} = \mathds{R}^n$, where $\mathcal{A}$ is a finite set of actions, $s_0$
	is the initial state and $P:\mathcal{B}(\mathds{R}^n)\times\mathcal{S}\times\mathcal{A}\rightarrow [0,1]$ is a Borel-measurable transition kernel which assigns to any state and any action a probability measure on the Borel space $(\mathds{R}^n,\mathcal{B}(\mathds{R}^n))$ \cite{stochastic}. 
	$\mathcal{AP}$ is a finite set of atomic propositions and a labelling
	function $L: \mathcal{S} \rightarrow 2^{\mathcal{AP}}$ assigns to each state
	$s \in \mathcal{S}$ a set of atomic propositions $L(s) \subseteq
	2^\mathcal{AP}$ \cite{stochastic}.$\hfill \lrcorner$
\end{definition}

Note that a finite-state MDP is a special case of continuous-state space MDP in which $|\mathcal{S}|<\infty$ and $P:\mathcal{S}\times\mathcal{A}\times\mathcal{S}\rightarrow[0,1]$ is defined as the transition probability function. The transition function $P$ induces a matrix which is usually known as transition probability matrix in the literature.

\begin{definition}[Path] 
In a continuous-state MDP $\textbf{M}$, an infinite path $\rho$ starting at $s_0$ is a sequence of states $\rho= s_0 \xrightarrow{a_0} s_1 \xrightarrow{a_1} ... ~$ such that every transition $s_i \xrightarrow{a_i} s_{i+1}$ is possible in $\textbf{M}$, i.e. $s_{i+1}$ belongs to the smallest Borel set $B$ such that $P(B|s_i,a_i)=1$. We might also denote $\rho$ as $s_0..$ to emphasize that $\rho$ starts from $s_0$.$\hfill\lrcorner$
\end{definition}

\begin{definition}
	[Stationary Policy] A stationary (randomized) policy $\mathit{Pol}: \mathcal{S} \times \mathcal{A} \rightarrow [0,1]$ is a mapping from each state $s \in \mathcal{S}$, and action $a \in \mathcal{A}$ to the probability of taking action $a$ in state $s$. A deterministic policy is a degenerate case of a randomized policy which outputs a single action at a given state, that is $\forall s \in \mathcal{S},~\exists a \in \mathcal{A},~\mathit{Pol}(s,a)=1$.$\hfill \lrcorner$
\end{definition}

In an MDP $\textbf{M}$, we define a function $R:\mathcal{S}\times\mathcal{A}\rightarrow \mathds{R}_0^+$ that denotes the immediate scalar bounded reward received by the agent from the environment after performing action $a \in \mathcal{A}$ in state $s \in \mathcal{S}$.

\begin{definition} [Expected (Infinite-Horizon) Discounted Reward] 
	\label{expectedut}
	For a policy $\mathit{Pol}$ on an MDP $\textbf{M}$, the expected discounted reward is defined as \cite{sutton}:
	\begin{equation}
	\label{expecteduteq}
	{U}^{\mathit{Pol}}(s)=\mathds{E}^{\mathit{Pol}} [\sum\limits_{n=0}^{\infty} \gamma^n~ R(s_n,Pol(s_n))|s_0=s],
	\end{equation}
	where $\mathds{E}^{\mathit{Pol}} [\cdot]$ denotes the expected value given that the agent follows policy $\mathit{Pol}$, $\gamma\in [0,1)$ is a discount factor and $s_0,...,s_n$ is the sequence of states generated by policy $\mathit{Pol}$ up to time step $n$. $\hfill \lrcorner$
\end{definition}


\begin{definition}[Optimal Policy]
	\label{optimal_policy}
	Optimal policy $\mathit{Pol}^*$ is defined as follows:
	$$
	\mathit{Pol}^*(s)=\arg\sup\limits_{Pol \in \mathcal{D}}~ {U}^{\mathit{Pol}}(s),
	$$
	where $\mathcal{D}$ is the set of all stationary deterministic policies over the state space $\mathcal{S}$.$\hfill \lrcorner$
\end{definition}

\begin{theorem}
	In any MDP $\textbf{M}$ with bounded reward function and finite action space, if there exists an optimal policy, then that policy is stationary and deterministic \cite{puterman} \cite{puterman2}.$\hfill\lrcorner$
\end{theorem}

An MDP $\textbf{M}$ is said to be solved if the agent discovers an optimal policy $\mathit{Pol}^*:\mathcal{S}\rightarrow\mathcal{A}$ to maximize the expected reward. From Definitions \ref{expectedut} and \ref{optimal_policy}, it means that the agent has to take actions that return the highest expected reward. In the following, we give background on Q-learning \cite{watkins}, which is probably the most widely used RL algorithm in solving finite-state MDPs. Later we present fundamentals of other approaches used in solving infinite-state MDPs.

\subsubsection{Classical Q-learning}

Let the MDP $\textbf{M}$ be a finite-state MDP. For each state $s \in \mathcal{S}$ and for any available action $a \in \mathcal{A}$, Q-learning (QL) assigns a quantitative measure through $Q:\mathcal{S}\times\mathcal{A}\rightarrow \mathds{R}$, which is initialized with an arbitrary finite value for all state-action pairs. During the learning, the Q-function is updated by the following rule when the agent takes action $ a $ at state $ s $:
\begin{equation}
\label{ql_update_rule}
\resizebox{0.91\columnwidth}{!}{
$Q(s,a) \leftarrow Q(s,a)+\mu
[R(s,a)+\gamma \max\limits_{a' \in \mathcal{A}}(Q(s',a'))-Q(s,a)],$
}
\end{equation}
where $ Q(s,a) $ is the Q-value corresponding to state-action $ (s,a) $, $
0<\mu\leq 1 $ is called learning rate (or step size), $ R(s,a) $ is the immediate reward obtained for performing action $a$ in state $s$, $\gamma$ is the 
discount factor, and $s'$ is the state obtained after performing action $a$. Q-function for the rest of the state-action pairs remains unchanged. 

Under mild assumptions, for finite-state and finite-action spaces QL converges to a unique limit, as long as every state action pair is visited infinitely often \cite{watkins}. Once QL converges, the optimal policy $\mathit{Pol}^*: \mathcal{S} \rightarrow \mathcal{A}$ can be generated by selecting the action that yields the highest $Q$, i.e.,
$$
\mathit{Pol}^*(s)=\arg\max\limits_{a \in \mathcal{A}}~Q(s,a),
$$
where $ \mathit{Pol}^* $ is the same optimal policy that can be generated via DP with Bellman operation. 

\subsubsection{Neural Fitted Q-iteration}

In the case when the MDP has a continuous state space the standard look-up table method of QL is practically infeasible to apply. Yet, we would like to preserve the valuable feature of QL, namely its independence from maintaining a \emph{model} and synthesising policies solely based on a set of experience samples. Neural Fitted Q-iteration (NFQ) \cite{nfq} achieves this by employing a multi-layer perceptron \cite{multilayer} to approximate the Q-function over the set of experience samples. NFQ, is the bare bone algorithm behind the ground-breaking algorithm, Deep Reinforcement Learning \cite{deepql}.

Following the same objective as QL and its update rule in \eqref{ql_update_rule}, NFQ defines a loss function that measures the error between the current Q-value and the new value that has to be assigned to the current Q-value: 
\begin{equation}
\label{loss_function}
\mathfrak{L}=(Q(s,a)-(R(s,a)+\gamma \max\limits_{a'}Q(s',a')))^2.
\end{equation}
Over this error, common gradient descent techniques can be applied to adjust the weights of the neural network, 
so that the error is minimized. 

In classical QL, the Q-function is updated whenever a state-action pair is visited. In the continuous state-space case, we may update the approximation in the same way, i.e., update the neural net weights once a new state-action pair is visited. However, in practice, a large number of trainings might need to be carried out until an optimal or near optimal policy is found. This is due to the uncontrollable changes occurring in the Q-function approximation caused by unpredictable changes in the network weights when the weights are adjusted for one certain state-action pair \cite{nfq-1}. More specifically, if at each iteration we only introduce a single sample point the training algorithm tries to adjust the weights of the neural network such that the loss function becomes minimum for that specific sample point. This might result in some changes in the network weights such that the error between the network output and the previous output of sample points becomes large and failure to approximate the Q-function correctly. 
Therefore, we have to make sure that when we update the weights of the neural network, we explicitly introduce previous samples as well: this technique is called ``experience replay" \cite{nfq-2} and detailed later. 

The core idea underlying NFQ is to gather experience samples and store them and then reuse this set every time the neural Q-function is updated. NFQ can be seen as a batch learning method in which there exists a training set that is repeatedly used to train the agent. In this sense NFQ is an offline algorithm as experience gathering and learning happens separately.

\subsubsection{Voronoi Quantizer}
\edit{As stated earlier, when the given MDP is continuous-state and the solution of choice is classical RL, the state space has to be discretised}. The discretisation can be done manually over the state space. However, one of the most appealing features of RL is its autonomy. In other words, RL is an unsupervised learning algorithm and therefore, the state space discretisation should be performed as part of the learning task, instead of being fixed at the start of the learning process.

Nearest neighbour vector quantization is a method for discretising the state space into a set of disjoint regions \cite{vectorq}. The Voronoi Quantizer (VQ) \cite{voronoi}, a nearest neighbour quantizer, maps the state space $\mathcal{S}$ onto a finite set of disjoint regions called Voronoi cells. The set of centroids of these cells is denoted by $\mathcal{C}=\{c_i\}_{i=1}^m,~c_i \in  \mathcal{S}$, where $m$ is the number of the cells. Therefore, designing a nearest neighbour vector quantizer deduces to coming up with the set $\mathcal{C}$. With $ \mathcal{C} $, we are able to use QL and find an approximation of the optimal policy for a continuous-state space MDP. The details of how the set of centroids $ \mathcal{C} $ is generated as part of the learning task is to follow.

\subsubsection{Fitted Value Iteration}
Finally, this section discusses Fitted Value Iteration (FVI) for continuous-state numerical DP using a function approximator \cite{gordon}. In standard value iteration the goal is to find a mapping (called value function) from the state space to $\mathds{R}$ such that it can lead the agent to find the optimal policy. The value function in our setup is (\ref{expecteduteq}) when $\mathit{Pol}$ is the optimal policy, i.e. ${U}^{\mathit{Pol}^*}$. In continuous state spaces, no analytical representation of the value function is in general available. Thus, an approximation can be obtained numerically through approximate value iteration, which involves approximately iterating the Bellman operator $T$ on some initial value function \cite{cdp}. FVI is explored later in the paper.

\section{Linear Temporal Logic Properties}
\label{LTL}

Ensuring that high-level requirements are accurately reflected in design elements of a learning algorithm is a key aspect of producing reliable policies in safety-critical problems. In this work, we employ Linear Temporal Logic (LTL) \cite{pnueli} as a high-level mission task to systematically shape the reward function in RL. An LTL formula can express a wide range of properties, such as safety and persistence. LTL formulas over a given set of atomic propositions $\mathcal{AP}$ are syntactically defined as
\begin{equation}
\label{ltlsyntax}
\varphi::= true ~|~ \alpha \in \mathcal{AP} ~|~ \varphi \land \varphi ~|~ \neg \varphi ~|~ \bigcirc \varphi ~|~ \varphi \cup \varphi.
\end{equation}
We define the semantics of LTL formula next, as interpreted over MDPs. Given a path $\rho$, the $i$-th state of $\rho$ is denoted by $\rho[i]$ where $\rho[i]=s_{i}$. Furthermore, the $i$-th suffix of $\rho$ is	$\rho[i..]$ where $\rho[i..]=s_i \xrightarrow{a_i} s_{i+1} \xrightarrow{a_{i+1}} s_{i+2} \xrightarrow{a_{i+2}} s_{i+3} \xrightarrow{a_{i+3}} ...~$.

\begin{definition}
	[LTL Semantics] \label{semantics} 
	For an LTL formula $\varphi$ and for a path $\rho$, the satisfaction relation $\rho\models\varphi$ is defined as
	\begin{equation*}
	\resizebox{0.98\columnwidth}{!}{$
	\begin{aligned}
	& \rho \models \alpha \in \mathcal{AP} \Leftrightarrow \alpha \in L(\rho[0]), \\
	& \rho \models \varphi_1\wedge \varphi_2 \Leftrightarrow \rho \models \varphi_1\wedge \rho \models \varphi_2,\\
	& \rho \models \neg \varphi \Leftrightarrow \rho \not \models \varphi, \\
	& \rho \models \bigcirc \varphi \Leftrightarrow \rho[1..] \models \varphi, \\
	& \rho \models \varphi_1\cup \varphi_2 \Leftrightarrow \exists j \geq 0 : \rho[j..] \models \varphi_2 \wedge\forall i, 0 \leq i < j, \rho[i..] \models \varphi_1.
	\end{aligned}$}
	\end{equation*} 
	\hfill $\lrcorner$
\end{definition}
Using the until operator we are able to define two temporal modalities: (1)
eventually, $\lozenge \varphi = true \cup \varphi$; and (2) always, $\square
\varphi = \neg \lozenge \neg \varphi$. \edit{LTL extends propositional logic with the temporal modalities until $ \cup $, eventually $ \lozenge $, and always $ \square $. For example, in a robot control problem, statements such as ``eventually get to this point'' or ``always stay safe'' are expressible by these modalities and can be combined via logical connectives and nesting to provide general and complex task specifications. Any LTL task specification} $\varphi$ over
$\mathcal{AP}$ expresses the following set of words:
$
\mathit{Words}(\varphi)=\{\sigma \in (2^{\mathcal{AP}})^\omega ~\mbox{s.t.}~ \sigma \models \varphi\}.
$
\begin{definition}[Policy Satisfaction]
	We say that a stationary deterministic policy $ \mathit{Pol} $ satisfies an LTL formula $ \varphi $ if
	$
	\mathds{P}[L(s_0)L(s_1)L(s_2)...\allowbreak\in \mathit{Words}(\varphi)]>0,
	$
	where every transition $s_i \rightarrow s_{i+1},~i=0,1,...$ is executed by taking action $ \mathit{Pol}(s_i) $ at state $ s_i $. \hfill $\lrcorner$ 
\end{definition}

%
An alternative method to express the set of associated
words, i.e., $Words(\varphi)$, is to employ a finite-state machine. Limit Deterministic B\"uchi Automatons (LDBA) \cite{sickert} are one of the most succinct finite-state machines for that purpose \cite{sickert2}. We need to first define a Generalized B\"uchi Automaton (GBA) and then we formally introduce an LDBA.

\begin{definition}\label{gba_definition}
	[Generalized B\"uchi Automaton] A GBA $\textbf{N}=(\allowbreak\mathcal{Q},\allowbreak q_0,\allowbreak\Sigma, \allowbreak\mathcal{F}, \allowbreak\Delta)$ is a structure where $\mathcal{Q}$ is a finite set of states, $q_0 \subseteq \mathcal{Q}$ is the set of initial states, $\Sigma=2^\mathcal{AP}$ is a finite alphabet, $\mathcal{F}=\{F_1,...,F_f\}$ is the set of accepting conditions where $F_j \subset \mathcal{Q}, 1\leq j\leq f$, and $\Delta: \mathcal{Q} \times \Sigma \rightarrow 2^\mathcal{Q}$ is a transition relation. \hfill $\lrcorner$ 
\end{definition}
Let $\Sigma^\omega$ be the set of all
infinite words over $\Sigma$. An infinite word $w \in \Sigma^\omega$ is
accepted by a GBA $\textbf{N}$ if there exists an infinite run $\theta \in
\mathcal{Q}^\omega$ starting from $q_0$ where $\theta[i+1] \in
\Delta(\theta[i],\omega[i]),~i \geq 0$ and for each $F_j \in \mathcal{F}$
\begin{equation} \label{acc}
\mathit{inf}(\theta) \cap F_j \neq \emptyset, 
\end{equation}
where $\mathit{inf}(\theta)$ is the set of states that are visited
infinitely often in the sequence $\theta$. 

\begin{definition}
	[LDBA]
	\label{ldbadef}
	A GBA $\textbf{N}=(\mathcal{Q},q_0,\Sigma, \mathcal{F}, \Delta)$ is
	limit deterministic if $\mathcal{Q}$ can be partitioned into two disjoint sets $\mathcal{Q}=\mathcal{Q}_N \cup \mathcal{Q}_D$, 
	such that \cite{sickert}:
	\begin{itemize}
		\item $\Delta(q,\alpha) \subseteq \mathcal{Q}_D$ and $|\Delta(q,\alpha)|=1$ for every state $q\in\mathcal{Q}_D$ and for every corresponding $\alpha \in \Sigma$,
		\item for every $F_j \in \mathcal{F}$, $F_j \subset \mathcal{Q}_D$. \hfill$\lrcorner$
	\end{itemize}
\end{definition}
An LDBA is a GBA that has two partitions: initial ($\mathcal{Q}_N$) and accepting ($\mathcal{Q}_D$). The accepting part includes all the accepting states and has deterministic transitions. Additionally, there are non-deterministic $\varepsilon$-transitions from $ \mathcal{Q}_N $ to $ \mathcal{Q}_D $. An $\varepsilon$-transition allows an automaton to change its state without reading an atomic proposition. 

%
%
\section{Logically-Constrained Neural Fitted Q-iteration}
\label{LCNFQ}

In this section, we propose an algorithm inspired by Neural Fitted Q-iteration (NFQ) that is able to synthesize a policy (or policies) that satisfies a temporal logic property. We call this algorithm Logically-Constrained NFQ (LCNFQ). We relate the notion of MDP and automaton by synchronizing them \emph{on-the-fly} to create a new structure that is first of all compatible with RL and second that embraces the logical property. 

\begin{definition} [Product MDP]
	\label{product_mdp_def} Given an MDP $\textbf{M}(\allowbreak\mathcal{S},\allowbreak\mathcal{A},\allowbreak s_0,\allowbreak P,\allowbreak\mathcal{AP},L)$ and
	an LDBA $\textbf{N}(\mathcal{Q},q_0,\Sigma, \mathcal{F}, \Delta)$ with
	$\Sigma=2^{\mathcal{AP}}$, the product MDP is defined as $(\textbf{M}\otimes
	\textbf{N}) = \textbf{M}_\textbf{N}=(\mathcal{S}^\otimes,\allowbreak
	\mathcal{A},\allowbreak s^\otimes_0,P^\otimes,\allowbreak
	\mathcal{AP}^\otimes,\allowbreak L^\otimes,\allowbreak
	\mathcal{F}^\otimes)$, where $\mathcal{S}^\otimes =
	\mathcal{S}\times\mathcal{Q}$, $s^\otimes_0=(s_0,q_0)$,
	$\mathcal{AP}^\otimes = \mathcal{Q}$, $L^\otimes =
	\mathcal{S}\times\mathcal{Q}\rightarrow 2^\mathcal{Q}$ such that
	$L^\otimes(s,q)={q}$ and $\mathcal{F}^\otimes
	\subseteq {\mathcal{S}^\otimes}$ is the set of accepting states $\mathcal{F}^\otimes=\{F^\otimes_1,...,F^\otimes_f\}$, where ${F}^\otimes_j=\mathcal{S}\times F_j$. 
	The transition kernel $P^\otimes$ is such that given the current state $(s_i,q_i)$ and action $a$, the new state is $(s_j,q_j)$, where $s_j\sim P(\cdot|s_i,a)$ and $q_j\in\Delta(q_i,L(s_j))$. Additionally, to handle $\varepsilon$-transitions the following modifications has to be added to the above definition of product MDP: \\*
	(1) for every potential $\varepsilon$-transition to some state $q \in \mathcal{Q}$ we add a corresponding action $\varepsilon_q$ in the product:
		
		$$
		\mathcal{A}^\otimes=\mathcal{A}\cup \{\varepsilon_q, q \in \mathcal{Q}\}.
		$$
		(2) the transition probabilities corresponding to $\varepsilon$- transitions are given by 
		\[P^\otimes((s_i,q_i),\varepsilon_q,(s_j,q_j)) = \left\{
		\begin{array}{lr}
		1 & $ if $  s_i=s_j,~q_i\xrightarrow{\varepsilon_q} q_j=q,\\
		0 & $ otherwise. $
		\end{array}
		\right.
		\]
	 \hfill $\lrcorner$
\end{definition}

By synchronising MDP states with LDBA states through the product MDP we add an extra dimension to the state space of the original MDP. The role of the added dimension is to track the automaton state and, hence, to synchronize the current state of the MDP with the state of the automaton and thus to evaluate the satisfaction of the associated LTL property. Note that LCNFQ is \emph{model-free} and there is no need to explicitly construct the product MDP. 

\begin{definition}
	[Accepting Frontier Function] \label{frontier} For an LDBA $\textbf{N}(\mathcal{Q},q_0,\Sigma,\allowbreak\mathcal{F},\allowbreak\Delta)$, we define the function $ Acc:\mathcal{Q}\times 2^{\mathcal{Q}}\rightarrow2^\mathcal{Q} $ as the accepting frontier function, which executes the following operation over a given set $ \mathds{F}\in 2^{\mathcal{Q}}$ 	
	\[Acc(q,\mathds{F})=\left\{
	\begin{array}{lr}
	\mathds{F}\setminus F_j~~~ & q \in F_j \wedge \mathds{F}\neq F_j,\\
	\\
	\bigcup\limits_{k=1}^{f} F_k \setminus F_j~~~ & q \in F_j \wedge \mathds{F}=F_j,\\
	\\
	\mathds{F} & $otherwise.$
	\end{array}
	\right. 
	\] 
	Once the state $ q\in F_j $ and the set $ \mathds{F} $ are introduced to the function $ Acc $, it outputs a set containing the elements of $ \mathds{F} $ minus those elements that are common with $ F_j $. 
	However, if $ \mathds{F}=F_j $, then the output is the union of all accepting sets of the LDBA minus those elements that are common with $ F_j $. Finally, if the state $ q $ is not an accepting state then $ Acc $ returns $ \mathds{F} $.	 \hfill $\lrcorner$
\end{definition}

The synchronised product MDP encompasses transition relations of the original MDP and the structure of the B\"uchi automaton, thus it inherits characteristics of both. Therefore, a proper reward function can lead RL to find a policy that is optimal with respect to satisfaction of the LTL property~$\varphi$. In this paper, we propose an on-the-fly reward function that observes the current state $ s^\otimes $, the action $ a $ and observes the subsequent state $ {s^\otimes}' $ and gives the agent a scalar value according to the following rule:
\begin{equation}\label{thereward}
\begin{aligned}
R(s^\otimes,a) = \left\{
\begin{array}{lr}
r_p & $ if $  q' \in \mathds{A},~{s^\otimes}'=(s',q'),\\
r_n & $ otherwise.$
\end{array}
\right.
\end{aligned}
\end{equation} 
Here $r_p=M+y\times m\times \mathit{rand}(s^\otimes)$ is a positive reward and $r_n=y \times m \times \mathit{rand}(s^\otimes)$ is a neutral reward. The parameter $y\in\{0,1\}$ is a constant, $0 < m \ll M$ are arbitrary positive values, and $\mathit{rand}: \mathcal{S}^\otimes \rightarrow (0,1)$ is a function that generates a random number in $(0,1)$ for each state $s^\otimes$ each time $ R $ is being evaluated. The role of the function $rand$ is to break the symmetry in LCNFQ, i.e. if all weights in a feedforward net \cite{apprx} start with equal values and if the solution requires that unequal weights be developed, the network can never learn. The reason is that the correlations between the weights within the same hidden layer can be described by symmetries in that layer, i.e. identical weights. Therefore, the neural net can generalize if such symmetries are broken and the redundancies of the weights are reduced. Starting with a completely identical weights prevents the neural net to minimize these redundancies and optimize the loss function. \edit{Also, note that parameter $y$ essentially acts as a switch to bypass the effect of the $rand$ function on $ R $. As we will see later, this switch is only active for LCNFQ.}

The set $ \mathds{A} $ is called the accepting frontier set, is initialised as $ \mathds{A}=\bigcup_{k=1}^{f} F_k $, 
and is updated by the following rule every time after the reward function is evaluated: 
$$
\mathds{A}\leftarrow Acc(q',\mathds{A}). 
$$
The set $ \mathds{A} $ always contains those accepting states that are needed to be visited at a given time. Thus, the agent is guided by the above reward assignment to visit these states and once all of the sets $ F_k,~k=1,...,f, $ are visited, the accepting frontier $ \mathds{A} $ is reset. As such, the agent is guided to visit the accepting sets infinitely often, and consequently, to satisfy the given LTL property (Definition \ref{gba_definition}). 


%
%
%

The LTL property is initially specified as a high-level formula $\varphi$ and is then converted to an LDBA $\textbf{N}$ to form a product MDP $\textbf{M}_\textbf{N}$ (see Definition \ref{product_mdp_def}). In order to use the experience replay technique we let the agent explore the MDP \edit{and reinitialize it when a positive reward is received or when no positive reward is received after $ \mathit{th} $ iterations. The parameter $ \mathit{th} $ is set manually according to the MDP and it allows the agent to explore the MDP and to prevent the sample set to explode in size.} All episode traces, i.e. experiences, are stored in the form of $(s^\otimes,a,{s^\otimes}',R(s^\otimes,a),q)$. Here $s^\otimes=(s,q)$ is the current state in the product MDP, $a$ is the chosen action, ${s^\otimes}'=(s',q')$ is the resulting state, and $R(s^\otimes,a)$ is the reward. The set of past experiences is called the sample set $\mathcal{E}$.

\edit{Once the exploration is finished and the sample set is created, we move forward to the learning phase. In the learning phase,} we employ $n$ separate \edit{multi-layer perceptrons with just one hidden layer} where $n=|\mathcal{Q}|$ and $\mathcal{Q}$ is the finite cardinality of the automaton $ \textbf{N} $\footnote{\edit{We have tried different embeddings such as one-hot encoding \cite{onehot} and integer encoding in order to approximate the global Q-function with a single feedforward net. However, we have observed poor performance since these encoding allows the network to assume an ordinal relationship between automaton states. Therefore, we have turned to the final solution of employing $n$ separate neural nets that work together in a hybrid manner to approximate the global Q-function.}}. 
Each neural net is associated with a state in the LDBA and together the neural nets approximate the Q-function in the product MDP. 
For each automaton state $q_i \in \mathcal{Q}$ the associated neural net is called $B_{q_i}:\mathcal{S}^\otimes\times\mathcal{A}\rightarrow\mathds{R}$. Once the agent is at state $s^\otimes=(s,q_i)$ the neural net $B_{q_i}$ is used for the local Q-function approximation. The set of neural nets acts as a global hybrid Q-function approximator $ Q:\mathcal{S}^\otimes\times\mathcal{A}\rightarrow\mathds{R} $. Note that the neural nets are not fully decoupled. For example, assume that by taking action $a$ in state $s^\otimes=(s,q_i)$ the agent is moved to state ${s^\otimes}'=(s',q_j)$ where $ q_i\neq q_j $. According to (\ref{loss_function}) the weights of $B_{q_i}$ are updated such that $B_{q_i}(s^\otimes,a)$ has minimum possible error to $R(s^\otimes,a)+\gamma\max_{a'} B_{q_j}({s^\otimes}',a')$. Therefore, the value of $B_{q_j}({s^\otimes}',a')$ affects $B_{q_i}(s^\otimes,a)$. 

Let $q_i \in \mathcal{Q}$ be a state in the LDBA. Then define $\mathcal{E}_{q_i}:=\{(\cdot,\cdot,\cdot,\cdot,x) \allowbreak \in  \mathcal{E} | x = q_i \}$ as the set of experiences within $\mathcal{E}$ that is associated with state $q_i$, i.e., $\mathcal{E}_{q_i}$ is the projection of $\mathcal{E}$ onto $q_i$. Once the experience set $\mathcal{E}$ is gathered, each neural net $B_{q_i}$ is trained by its associated experience set $\mathcal{E}_{q_i}$. At each iteration a pattern set $\mathcal{P}_{q_i}$ is generated based on $\mathcal{E}_{q_i}$:
$$
\mathcal{P}_{q_i}=\{(input_l,target_l), l=1,...,|\mathcal{E}_{q_i}|)\},$$
where $input_l=({s_l}^\otimes,a_l)$ and $target_l=\allowbreak R({s_l}^\otimes,a_l)+\allowbreak\gamma \max_{a'\in\mathcal{A}}\allowbreak Q({{s_l}^\otimes}',a')$ such that $({s_l}^\otimes,a_l,{{s_l}^\otimes}',R({s_l}^\otimes,a_l),{q_i}) \in \mathcal{E}_{q_i}
$. The pattern set is used to train the neural net $B_{q_i}$. We use Rprop \cite{rprop} to update the weights in each neural net, as it is known to be a fast and efficient method for batch learning \cite{nfq}. In each cycle of LCNFQ (Algorithm \ref{lcnfqal}), the training schedule starts from networks that are associated with accepting states of the automaton and goes backward until it reaches the networks that are associated to the initial states. In this way we allow the Q-value to back-propagate through the networks. LCNFQ stops when the generated policy stops improving for long enough.

\begin{algorithm2e}[!t]
	\DontPrintSemicolon
	\SetKw{return}{return}
	\SetKwRepeat{Do}{do}{while}
	\SetKwData{conflict}{conflict}
	\SetKwData{safe}{safe}
	\SetKwData{sat}{sat}
	\SetKwData{unsafe}{unsafe}
	\SetKwData{unknown}{unknown}
	\SetKwData{true}{true}
	\SetKwInOut{Input}{input}
	\SetKwInOut{Output}{output}
	\SetKwFor{Loop}{Loop}{}{}
	\SetKw{KwNot}{not}
	\begin{small}
		\Input{MDP $\textbf{M}$, a set of transition samples $\mathcal{E}$}
		\Output{Approximated Q-function}
		initialize all neural nets $B_{q_i}$ with $ (s_0,q_i,a) $ as the input and $ r_n $ as the output where $ a\in\mathcal{A} $ is a random action\;
		\Repeat{end of trial}
		{
			\For{$q_i=|\mathcal{Q}|$ \textbf{to} $1$}
			{
				$\mathcal{P}_{q_i}=\{(input_l,target_l),~l=1,...,|\mathcal{E}_{q_i}|)\}$\;
				~~~~~~~~~$input_l=({s_l}^\otimes,a_l)$\;
				~~~~~~~~~~\hspace{1mm}$target_l=R({s_l}^\otimes,a_l)+\gamma \max \limits_{a'} Q({{s_l}^\otimes}',a')$\;
				~~~~~~~~~where $({s_l}^\otimes,a_l,{{s_l}^\otimes}',R({s_l}^\otimes,a_l),{q_i}) \in \mathcal{E}_{q_i}$\;
				$B_{q_i} \leftarrow$ Rprop$(\mathcal{P}_{q_i})$
			}
		}
	\end{small}
	\caption{LCNFQ}
	\label{lcnfqal}
\end{algorithm2e}


Recall that the reward function \eqref{thereward} only returns a positive value when the agent has a transition to an accepting state in the product MDP. Therefore, if accepting states are reachable, by following this reward function the agent is able to come up with a policy ${\mathit{Pol}^\otimes}^*$ that leads to the accepting states. This means that the trace of read labels over $\mathcal{S}$ (see Definition \ref{product_mdp_def}) results in an automaton state to be accepting. Therefore, the trace over the original MDP is a trace that satisfies the given logical property. Also, recall that the optimal policy has the highest expected reward comparing to other policies. Consequently, the optimal policy has the highest expected probability of reaching to the accepting set, i.e. satisfying the LTL property. 

\edit{The next section studies state-space discretisation as the most popular alternative approach to solving infinite-state MDPs.}

\section{Voronoi Quantizer}
\label{vorosection}

Inspired by \cite{voronoi}, we propose a version of Voronoi quantizer that is able to discretise the state space of the product MDP $\mathcal{S}^\otimes$. In the beginning, $\mathcal{C}$ is initialized to consist of just one $c_1$, which corresponds to the initial state. This means that the agent views the entire state space as a
homogeneous region when no a priori knowledge is available. Subsequently, when the agent explores, the Euclidean distance between each newly visited state and its nearest neighbour is calculated. If this distance is greater than a threshold value $\Delta$ called \emph{minimum resolution}, or if the new state $ s^\otimes $ has a never-visited automaton state then the newly visited state is appended to $\mathcal{C}$. Therefore, as the agent continues to explore, the size of $\mathcal{C}$ would increase until the \emph{relevant} parts of the state space are partitioned. In our algorithm, the set $\mathcal{C}$ has $n$ disjoint subsets where $n=|\mathcal{Q}|$ and $\mathcal{Q}$ is the finite set of states of the automaton. Each subset $\mathcal{C}^{q_j},~j=1,...,n$ contains the centroids of those Voronoi cells that have the form of $c_i^{q_j}=(\cdot,q_j)$, i.e. $ \bigcup_i^m c_i^{q_j} = \mathcal{C}^{q_j}$ and $ \mathcal{C}=\bigcup_{j=1}^n \mathcal{C}^{q_j}$. Therefore, a Voronoi cell
$$
\{(s,q_j) \in \mathcal{S}^\otimes,||(s,q_j)-c_i^{q_j}||_2\leq||(s,q_j)-c_{i'}^{q_j}||_2 \},
$$
is defined by the nearest
neighbour rule for any $i'\neq i$. The VQ algorithm is presented in Algorithm \ref{voronoi}. The proposed algorithm consists of several resets at which the agent is forced to re-localize to its initial state $ s_0 $. Each reset is called an episode, as such in the rest of the paper we call this algorithm episodic VQ.

\begin{algorithm2e}[!t]
	\DontPrintSemicolon
	\SetKw{return}{return}
	\SetKwRepeat{Do}{do}{while}
	\SetKwData{conflict}{conflict}
	\SetKwData{safe}{safe}
	\SetKwData{sat}{sat}
	\SetKwData{unsafe}{unsafe}
	\SetKwData{unknown}{unknown}
	\SetKwData{true}{true}
	\SetKwInOut{Input}{input}
	\SetKwInOut{Output}{output}
	\SetKwFor{Loop}{Loop}{}{}
	\SetKw{KwNot}{not}
	\begin{small}
		\Input{MDP $\textbf{M}$, minimum resolution $\Delta$}
		\Output{Approximated Q-function $Q$}
		initialize $Q(c_1,a)=0,~\forall a \in \mathcal{A}$\;
		\Repeat{end of trial}
		{
			initialize $c_1=$ initial state\;
			set $c=c_1$\;
			\Repeat{end of trial}
			{
				$\alpha=\arg\max_{a\in\mathcal{A}} Q(c,a)$\;
				execute action $\alpha$ and observe the next state $(s',q)$\;
				\uIf{$\mathcal{C}_{q}$ is empty}
				{
					append $c_{\mathit{new}}=(s',q)$ to $\mathcal{C}_{q}$\;
					initialize $Q(c_{\mathit{new}},a)=0,~\forall a \in \mathcal{A}$\;
				}
				\Else{
					determine the nearest neighbor $c_{\mathit{new}}$ within $\mathcal{C}_q$\;
					\uIf{$||c_{\mathit{new}}-(s',q)||_2>\Delta$}
					{
					append $c_{\mathit{new}}=(s',q)$ to $\mathcal{C}_{q}$\;
					initialize $Q(c_{\mathit{new}},a)=0,~\forall a \in \mathcal{A}$\;
					}
					\Else{
						
						$Q(c,\alpha)=(1-\mu)Q(c,\alpha)+\mu
						[R(c,\alpha)+\gamma \max\limits_{a'}(Q(c_{\mathit{new}},a'))]$
					}
				}
				$c=c_{\mathit{new}}$\;
			}
		}
	\end{small}
	\caption{Episodic VQ}
	\label{voronoi}
\end{algorithm2e}

\section{Fitted Value Iteration}
\label{FVIsection}

In this section we propose a modified version of FVI that can handle the product MDP. The global value function $v:\mathcal{S}^\otimes\rightarrow\mathds{R}$, or more specifically $v:\mathcal{S}\times\mathcal{Q}\rightarrow\mathds{R}$, consists of $n$ number of sub-value functions where $n=|\mathcal{Q}|$. For each $ q_j\in\mathcal{Q} $, the sub-value function $v^{q_j}:\mathcal{S}\rightarrow\mathds{R}$ returns the value the states of the form $(s,q_j)$. In the same manner as LCNFQ, the sub-value functions are not decoupled. 

Let $P^\otimes(dy|s^\otimes,a)$ be the distribution over $\mathcal{S}^\otimes$ for the successive state given that the current state is $s^\otimes$ and the current action is $a$. For each state $ (s,q_j) $, the Bellman update over each sub-value function $v^{q_j}$ is defined as:
\begin{equation}
\label{bellman}
Tv^{q_j}(s)=\sup\limits_{a\in\mathcal{A}} \{\int v(y) P^\otimes(dy|(s,q_j),a)\},
\end{equation}
where $T$ is the Bellman operator \cite{hernandez}. The update in \eqref{bellman} is a special case of general Bellman update as it does not have a running reward and the (terminal) reward is embedded via value function initialization. The value function is initialized according to the following rule:
\begin{equation}
\label{fviinit}
v(s^\otimes) = \left\{
\begin{array}{lr}
r_p & $ if $ {s^\otimes} \in \mathbb{A},\\
r_n & $ otherwise. 
$
\end{array}
\right.
\end{equation}


where $ r_p $ and $ r_n $ are defined in \eqref{thereward}. The main hurdle in executing the Bellman operator in continuous state MDPs, as in (\ref{bellman}), is that no analytical representation of the value function $v$ and also sub-value functions $ v^{q_j},~q_j\in\mathcal{Q} $ is available. Therefore, we employ an approximation method by introducing the operator $L$. The operator $L$ constructs an approximation of the value function denoted by $ Lv $ and of each sub-value function $v^{q_j}$ which we denote by $Lv^{q_j}$. For each $ q_j\in\mathcal{Q} $ the approximation is based on a set of points $\{(s_i,q_j)\}_{i=1}^k \subset \mathcal{S}^\otimes$ which are called centres. For each $ q_j $, the centres $ i=1,...,k $ are distributed uniformly over $ \mathcal{S} $ such that they uniformly cover $ \mathcal{S} $.

We employ a kernel-based approximator for our FVI algorithm. Kernel-based approximators have attracted a lot of attention mostly because they perform very well in high-dimensional state spaces \cite{cdp}. One of these methods is the kernel averager, which can be represented by the following expression for each state $ (s,q_j) $:
\begin{equation}
\label{kernel}
Lv(s,q_j)=Lv^{q_j}(s)=\dfrac{\sum_{i=1}^{k} K(s_i-s) v^{q_j}(s_i)}{\sum_{i=1}^{k} K(s_i-s)},
\end{equation}
where the kernel $K:\mathcal{S}\rightarrow\mathds{R}$ is a radial basis function, such as $e^{-|s-s_i|/h}$, and $h$ is smoothing parameter. Each kernel has a centre $s_i$ and the value of it decays to zero as $s$ diverges from $s_i$. This means that for each $ q_j\in\mathcal{Q} $ the approximation operator $L$ in (\ref{kernel}) is a convex combination of the values of the centres $\{s_i\}_{i=1}^{k}$ with larger weight given to those values $v^{q_j}(s_i)$ for which $s_i$ is close to $s$. Note that the smoothing parameter $h$ controls the weight assigned to more distant values. 


\begin{algorithm2e}[!t]
	\DontPrintSemicolon
	\SetKw{return}{return}
	\SetKwRepeat{Do}{do}{while}
	\SetKwData{conflict}{conflict}
	\SetKwData{safe}{safe}
	\SetKwData{sat}{sat}
	\SetKwData{unsafe}{unsafe}
	\SetKwData{unknown}{unknown}
	\SetKwData{true}{true}
	\SetKwInOut{Input}{input}
	\SetKwInOut{Output}{output}
	\SetKwFor{Loop}{Loop}{}{}
	\SetKw{KwNot}{not}
	\begin{small}
		\Input{MDP $\textbf{M}$, a set of samples $\{s^\otimes_i\}_{i=1}^k=\{(s_i,q_j)\}_{i=1}^k$ for each $q_j\in\mathcal{Q}$, Monte Carlo sampling number $Z$, smoothing parameter $h$}
		\Output{approximated value function $Lv$}
		initialize $Lv$ \;
		sample $\mathcal{Y}_a^Z(s_i,q_j),~\forall q_j \in \mathcal{Q},~\forall i=1,...,k~,~\forall a\in\mathcal{A}$\;
		\Repeat{end of trial}
		{
			\For{$j=|\mathcal{Q}|$ \textbf{to} $1$}
			{
				$~\forall q_j \in \mathcal{Q},~\forall i=1,...,k~,~\forall a\in\mathcal{A}$ calculate $I_a((s_i,q_j))=1/Z \sum_{y \in \mathcal{Y}_a^Z(s_i,q_j)} Lv(y)~~~~~$ using \eqref{kernel}\;
				for each state $(s_i,q_j)$, update $v^{q_j}(s_i)=\sup_{a\in\mathcal{A}}\{I_a((s_i,q_j))\}$ in  \eqref{kernel}
			}
		}
	\end{small}
	\caption{FVI}
	\label{FVI}
\end{algorithm2e}

In order to approximate the integral in the Bellman update (\ref{bellman}) we use a Monte Carlo sampling technique \cite{montec}. For each center $(s_i,q_j)$ and for each action $a$, we sample the next state $y_a^z(s_i,q_j)$ for $z=1,...,Z$ times and append it to set of $ Z $ subsequent states $\mathcal{Y}_a^Z(s_i,q_j)$. We then replace the integral with 
\begin{equation}
\label{montec}
I_a(s_i,q_j)=\dfrac{1}{Z} \sum\limits_{z=1}^{Z} Lv(y_a^z(s_i,q_j)).
\end{equation}

The approximate value function $ Lv $ is initialized according to \eqref{fviinit}. In each cycle of FVI, the approximate Bellman update is first performed over the sub-value functions that are associated with accepting states of the automaton, i.e. those that have initial value of $ r_p $, and then goes backward until it reaches the sub-value functions that are associated to the initial states. In this manner, we allow the state values to back-propagate through the transitions that connects the sub-value function via \eqref{montec}. Once we have the approximated value function, we can generate the optimal policy by following the maximum value (Algorithm \ref{FVI}). 


\begin{figure}[!t]
	\centering
	\subfloat[Melas Chasma]{{\includegraphics[width=0.2\textwidth]{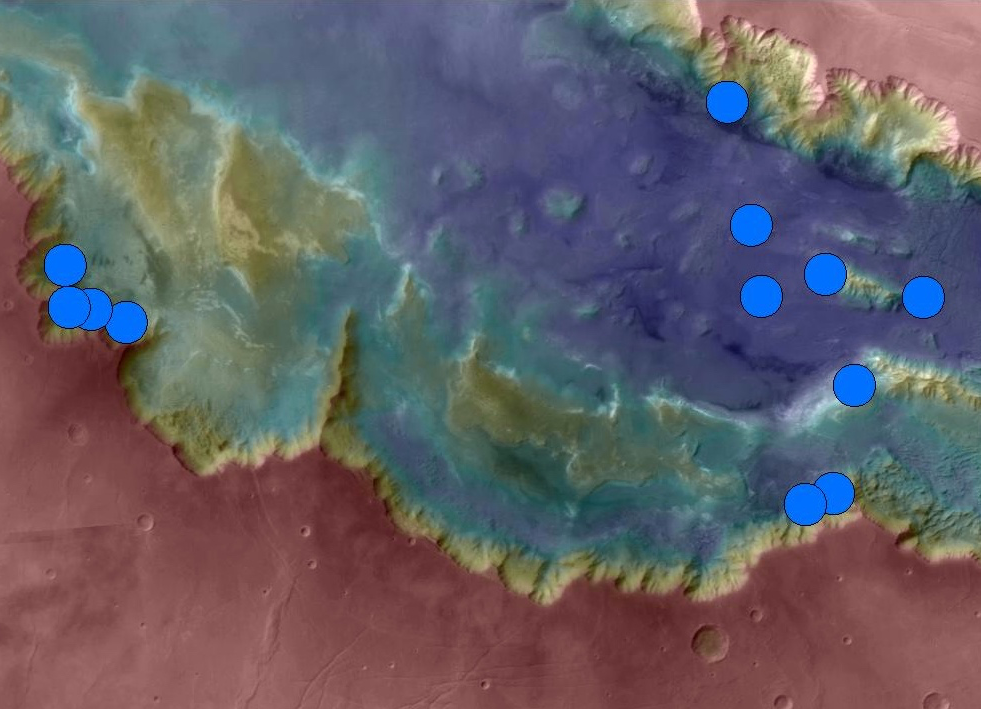} }}%
	\qquad
	\subfloat[Coprates Chasma]{{\includegraphics[width=0.21\textwidth]{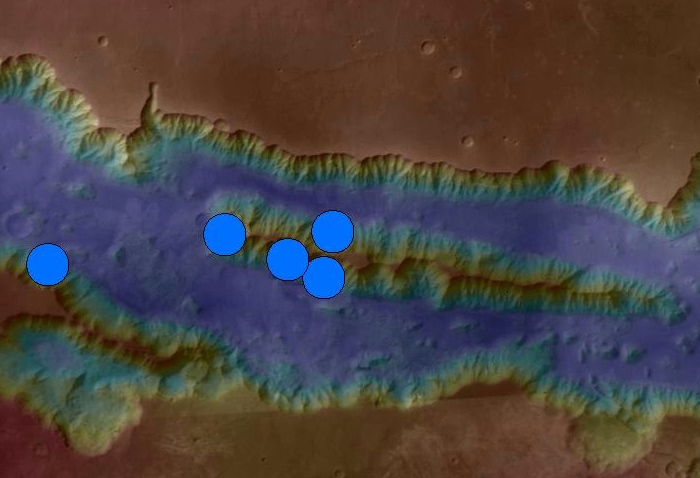} }}%
	\caption{Melas Chasma and Coprates Chasma, in the central and eastern portions of Valles Marineris. Map color spectrum represents elevation, where red is high and blue is low. (Image courtesy of NASA, JPL, Caltech and University of Arizona.)}%
	\label{MDPs}%
\end{figure}

\section{Experiments}
\label{case study}

We describe a mission planning task for an autonomous Mars rover that uses LCNFQ to pursue an exploration mission. The scenario of interest is that we start with an image from the surface of Mars and then we add the desired labels from $ 2^\mathcal{AP} $, e.g. safe or unsafe, to the image. We assume that we know the highest possible disturbance caused by different factors (such as sand storms) on the rover motion. This assumption can be set to be very conservative given the fact that there might be some unforeseen factors that we did not take into account.  

The next step is to express the desired mission in LTL format and run LCNFQ on the labelled image before sending the rover to Mars. We would like the rover to satisfy the given LTL property with the highest probability possible starting from any random initial state (as we can not predict the landing location exactly). Once LCNFQ is trained we use the network to guide the rover on the Mars surface.
We compare LCNFQ with Voronoi quantizer and FVI and we show that LCNFQ outperforms these methods.

\subsection{MDP Structure}
In this numerical experiment the area of interest on Mars is Coprates quadrangle, which is named after the Coprates River in ancient Persia. There exist a significant number of signs of water, with ancient river valleys and networks of stream channels showing up as sinuous and meandering ridges and lakes. We consider two parts of Valles Marineris, a canyon system in Coprates quadrangle (Fig. \ref{MDPs}). The blue dots, provided by NASA, indicate locations of recurring slope lineae (RSL) in the canyon network. RSL are seasonal dark streaks regarded as the strongest evidence for the possibility of liquid water on the surface of Mars. RSL extend down-slope during a warm season and then disappear in the colder part of the Martian year \cite{water_on_mars}. The two areas mapped in Fig. \ref{MDPs}, Melas Chasma and Coprates Chasma, have the highest density of known RSL. 

For each case, let the entire area be our MDP state space $\mathcal{S}$, where the rover location is a single state $s \in \mathcal{S} $. At each state $s \in \mathcal{S}$, the rover has a set of actions $ \mathcal{A}=\{\mathit{left},\mathit{right},\mathit{up},\mathit{down},\mathit{stay}\}$ by which it is able to move to other states: at each state $s \in \mathcal{S}$, when the rover takes an action $a \in \{\mathit{left},\mathit{right},\mathit{up},\mathit{down}\}$ it is moved to another state (e.g., $s'$) towards the direction of the action with a range of movement that is randomly drawn from $(0,D]$ unless the rover hits the boundary of the area which forces the rover to remain on the boundary. In the case when the rover chooses action $a=\mathit{stay}$ it is again moved to a random place within a circle centred at its current state and with radius $d \ll D$. Again, $ d $ captures disturbances on the surface of Mars and can be tuned accordingly. 

With $\mathcal{S}$ and $\mathcal{A}$ defined we are only left with the labelling function $L:\mathcal{S}\rightarrow 2^{\mathcal{AP}}$ which assigns to each state $s \in \mathcal{S}$ a set of atomic propositions $L(s) \subseteq 2^{\mathcal{AP}}$. With the labelling function, we are able to divide the area into different regions and define a logical property over the traces that the agent generates. In this particular experiment, we divide areas into three main regions: neutral, unsafe and target. The target label goes on RSL (blue dots), the unsafe label lays on the parts with very high elevation (red coloured) and the rest is neutral. In this example we assume that the labels do not overlap each other.

Note that when the rover is deployed to its real mission, the precise landing location is not known. Therefore, we should take into account the randomness of the initial state $ s_0 $. The dimensions of the area of interest in Fig. \ref{MDPs}.a are $ 456.98\times 322.58 $ km and in Fig. \ref{MDPs}.b are $ 323.47 \times 215.05 $ km. The diameter of each RSL is $ 19.12 $ km. Other parameters in this numerical example have been set as $D=2$ km, $d=0.02$ km, the reward function parameter $y=1$ for LCNFQ and $y=0$ for VQ and FVI, $M=1$, $m=0.05$ and $\mathcal{AP}=\{$neutral, unsafe, target$\_$1, target$\_$2$\}$.

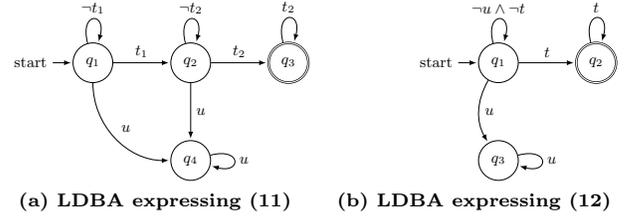
\begin{figure}[!t]
	\centering
	\subfloat[LDBA expressing (\ref{omega})]{{\hspace{0cm}
			\scalebox{.65}{
				\begin{tikzpicture}[shorten >=1pt,node distance=2cm,on grid,auto] 
				\node[state,initial] (q_1)   {$q_1$}; 
				\node[state] (q_2) [right=of q_1] {$q_2$}; 
				\node[state] (q_5) [below=of q_2] {$q_4$}; 
				\node[state,accepting] (q_3) [right=of q_2] {$q_3$}; 
				\path[->] 
				(q_1) edge [loop above] node {$\neg t_1$} ()   	
				(q_1) edge  node {$t_1$} (q_2)
				(q_2) edge [loop above] node {$\neg t_2$} ()
				(q_2) edge node {$t_2$} (q_3)
				(q_3) edge [loop above] node {$t_2$} () 
				(q_1) edge [bend right=45] node {$u$} (q_5)
				(q_2) edge node {$u$} (q_5)
				(q_5) edge [loop right] node {$u$} ();
				\end{tikzpicture}} }}%
	\qquad~~~~~
	\subfloat[LDBA expressing (\ref{ltl})]{{\hspace{1cm}
			\scalebox{0.65}{
				\begin{tikzpicture}[shorten >=1pt,node distance=2cm,on grid,auto] 
				\node[state,initial] (q_1)   {$q_1$}; 
				\node[state,accepting] (q_2) [right=of q_1] {$q_2$}; 
				\node[state] (q_3) [below=of q_1] {$q_3$}; 
				\path[->] 
				(q_1) edge  node {$t$} (q_2)
				(q_1) edge [bend right] node {$u$} (q_3)
				edge [loop above] node {$\neg u \wedge \neg t$} ()
				(q_2) edge [loop above] node {$t$} ()
				(q_3) edge [loop right] node {$u$} ();
				\end{tikzpicture}} }}%
	\caption{Generated LDBAs}%
	\label{f3}%
\end{figure}

\begin{figure}[!t]
	\centering
	\subfloat[Melas Chasma and landing location (black rectangle) $ (118,~85) $]{{\includegraphics[width=\x\textwidth]{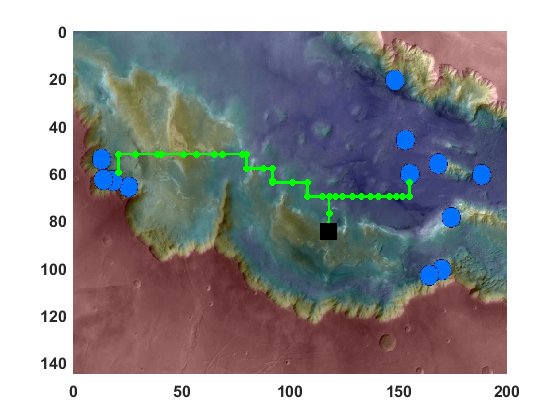}}}
	\qquad
	\subfloat[Coprates Chasma and landing location (black rectangle) $ (194,~74) $]{{\includegraphics[width=\x\textwidth]{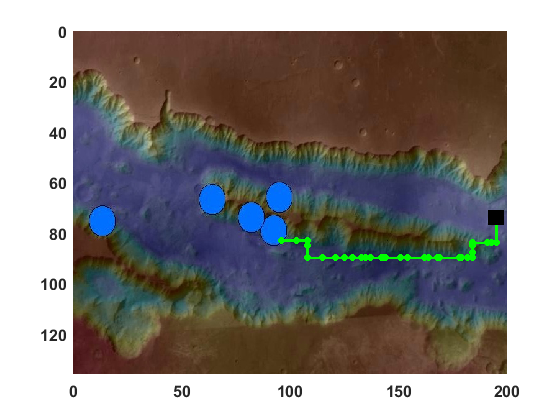} }}%
	\caption{\edit{Generated paths by} LCNFQ}
	\label{MDPs_solved}
\end{figure}
\begin{figure}[!t]
	\centering
	\subfloat[Melas Chasma and landing location (black rectangle) $ (118,~85) $]{{\includegraphics[width=\x\textwidth]{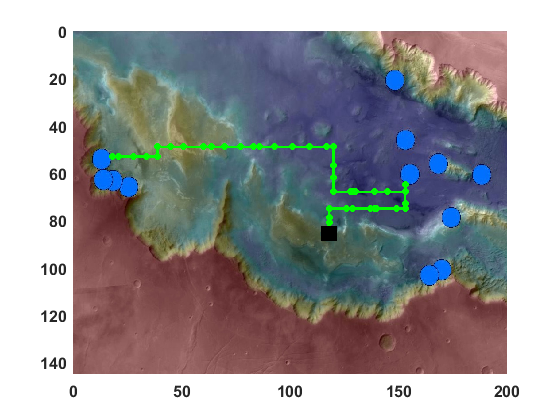}}}
	\qquad
	\subfloat[Coprates Chasma and landing location (black rectangle) $ (194,~74) $]{{\includegraphics[width=\x\textwidth]{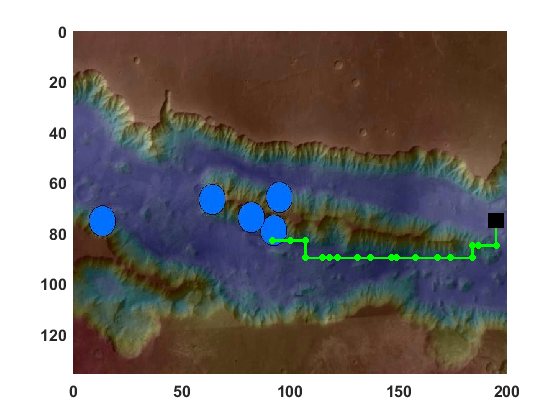} }}%
	\caption{\edit{Generated paths by} episodic VQ}
	\label{MDPs_solved_vq}
\end{figure}

\begin{figure}[!t]
	\centering
	\subfloat[Melas Chasma and landing location (black rectangle) $ (118,~85) $]{{\includegraphics[width=\x\textwidth]{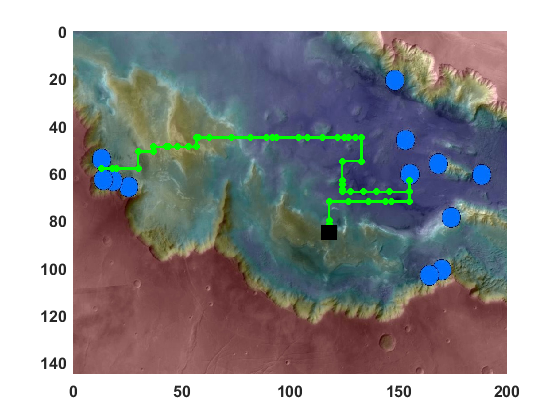}}}
	\qquad
	\subfloat[Coprates Chasma and landing location (black rectangle) $ (194,~74) $]{{\includegraphics[width=\x\textwidth]{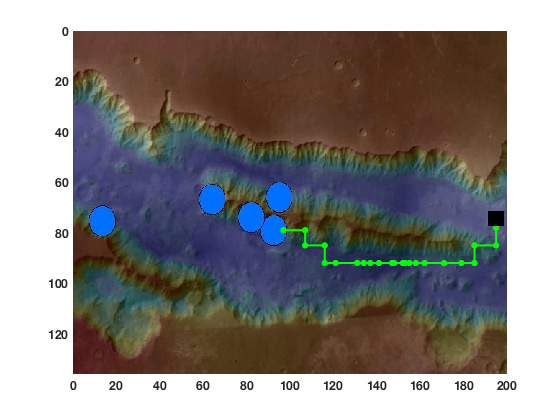} }}%
	\caption{\edit{Generated paths by} FVI}
	\label{MDPs_solved_fvi}
\end{figure}

\subsection{Specifications}

The first control objective in this numerical example is expressed by the following LTL formula over Melas Chasma (Fig. \ref{MDPs}.a):
\begin{equation}
\label{omega}
\lozenge(t_1 \wedge \lozenge t_2) \wedge \square(t_2 \rightarrow \square t_2) \wedge \square(u \rightarrow \square u),   
\end{equation} 
where $n$ stands for ``neutral", $t_1$ stands for ``target 1", $t_2$ stands for ``target 2" and $u$ stands for ``unsafe". Target 1 are the RSL (blue bots) on the right with a lower risk of the rover going to unsafe region and the target 2 label goes on the left RSL that are a bit riskier to explore. Conforming to (\ref{omega}) the rover has to visit the target 1 (any of the right dots) at least once and then proceed to the target 2 (left dots) while avoiding unsafe areas. Note that according to $ \square(u \rightarrow \square u) $ in \eqref{omega} the agent is able to go to unsafe area $ u $ (by climbing up the slope) but it is not able to come back due to the risk of falling. With (\ref{omega}) we can build the associated B\"uchi automaton as in Fig.~\ref{f3}.a. 

The second formula focuses more on safety and we are going to employ it in exploring Coprates Chasma (Fig. \ref{MDPs}.b) where a critical unsafe slope exists in the middle of this region: 	
\begin{equation}
\label{ltl}
\lozenge t \wedge \square(t \rightarrow \square t) \wedge \square(u \rightarrow \square u)   
\end{equation}

In \eqref{ltl}, $t$ refers to ``target", i.e. RSL in the map, and $u$ stands for ``unsafe". According to this LTL formula, the agent has to eventually reach the target ($\lozenge t$) and stays there ($\square(t \rightarrow \square t)$). However, if the agent hits the unsafe area it can never comes back and remains there forever ($\square(u \rightarrow \square u)$). With (\ref{ltl}) we can build the associated B\"uchi automaton as in Fig.~\ref{f3}.b. Having the B\"uchi automaton for each formula, we are able to use Definition~\ref{product_mdp_def} to build product MDPs and run LCNFQ on both.  

\begin{table}[!t]
	\caption{Simulation results}
	\label{table:1}
	\centering
	\renewcommand{\arraystretch}{1.5}
	\scalebox{0.5}{
		\begin{tabular}{|c|c|c|c|c|c|}
			\hline
			\multicolumn{6}{|c|}{Melas Chasma} \\
			\hline
			\textbf{Algorithm} & \textbf{Sample Complexity} & \textbf{\boldmath$ {U}^{\mathit{Pol}^*}(s_0) $\unboldmath} & \textbf{Success Rate$^\dag$} & \textbf{Training Time$^*$(s)} & \textbf{Iteration Num.}\\
			\hline
			LCNFQ & \textbf{7168} samples & \textbf{0.0203} & \textbf{99\%} & 95.64 & \textbf{40}\\
			\hline
			VQ ($\Delta=0.4$) & 27886 samples & 0.0015 & \textbf{99\%} & 1732.35 & 2195\\
			\hline
			VQ ($\Delta=1.2$) & 7996 samples & 0.0104 & 97\% & 273.049 & 913\\
			\hline
			VQ ($\Delta=2$) & - & 0 & 0\% & - & -\\
			\hline
			FVI & 40000 samples & 0.0133 & 98\% & \textbf{4.12} & 80\\
			\hline
			\multicolumn{6}{|c|}{Coprates Chasma} \\
			\hline
			\textbf{Algorithm} & \textbf{Sample Complexity} & \textbf{\boldmath$ {U}^{\mathit{Pol}^*}(s_0) $\unboldmath} & \textbf{Success Rate$^\dag$} & \textbf{Training Time$^*$(s)} & \textbf{Iteration Num.}\\
			\hline
			LCNFQ & \textbf{2680 samples} & \textbf{0.1094} & \textbf{98\%} & 166.13 & \textbf{40}\\
			\hline
			VQ ($\Delta=0.4$) & 8040 samples & 0.0082 & \textbf{98\%} & 3666.18 & 3870\\
			\hline
			VQ ($\Delta=1.2$) & 3140 samples & 0.0562 & 96\% & 931.33 & 2778\\
			\hline
			VQ ($\Delta=2$) & - & 0 & 0\% & - & -\\
			\hline
			FVI & 25000 samples & 0.0717 & 97\% & \textbf{2.16} & 80\\
			\hline
	\end{tabular}} 
	\renewcommand{\arraystretch}{1}
	~\\
	$ {}^\dag $ Testing the trained agent (for 100 trials) \\ 
	$ ~~~~~~{}^* $ Average for 10 trainings
\end{table}

\subsection{Simulation Results}

This section presents the simulation results. All simulations are carried on a machine with a 3.2GHz Core i5 processor and 8GB of RAM, running Windows 7. LCNFQ has four feedforward neural networks for (\ref{omega}) and three feedforward neural networks for (\ref{ltl}), each associated with an automaton state in Fig.~\ref{f3}.a and Fig.~\ref{f3}.b. We assume that the rover lands on a random safe place and has to find its way to satisfy the given property in the face of uncertainty. The learning discount factor $\gamma$ is also set to be equal to $0.9$. 

Fig.~\ref{MDPs_solved} gives the results of learning for the LTL formulas in (\ref{omega}) and (\ref{ltl}). At each state $s^\otimes$, the robot picks an action that yields highest $Q(s^\otimes,\cdot)$ and by doing so the robot is able to generate a control policy ${\mathit{Pol}^\otimes}^*$ over the state space $\mathcal{S}^\otimes$.  The control policy ${\mathit{Pol}^\otimes}^*$ induces a policy $\mathit{Pol}^*$ over the state space $\mathcal{S}$ and its performance is shown in Fig.~\ref{MDPs_solved}.

\begin{figure}[!t] \centering \includegraphics[width=0.9\columnwidth]{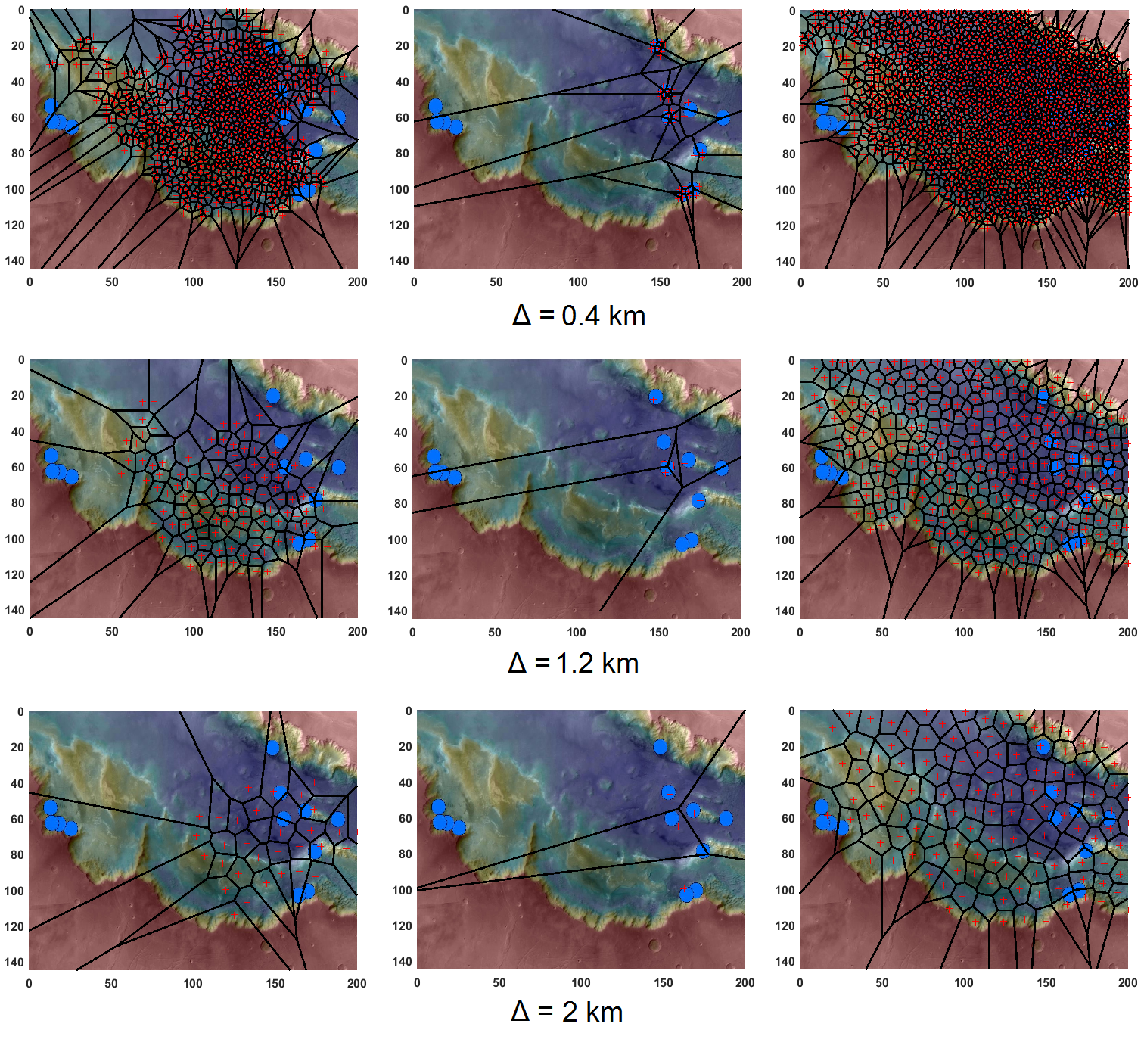} \caption{VQ generated cells in Melas Chasma for different resolutions. Note that the quantizer only focused on parts of the state space that are relevant to satisfaction of the LTL property}
	\label{vq_cells} 
\end{figure} 

Next, we investigate the episodic VQ algorithm as an alternative solution to LCNFQ. Three different resolutions ($\Delta=0.4,~1.2,~2$ km) are compared against the quality of the generated policy. The results are presented in Table \ref{table:1}, where VQ with $\Delta=2$ km fails to find a satisfying policy in both regions, due to the coarseness of the resulted discretisation. A coarse partitioning results in the RL not to be able to efficiently back-propagate the reward or the agent to be stuck in some random-action loop as sometimes the agent's current cell is large enough that all actions have the same value. In Table \ref{table:1}, training time is the empirical time that is taken to train the algorithm and travel distance is the distance that agent traverses from initial state to final state. We show the generated policy for $\Delta=1.2$ km in Fig. \ref{MDPs_solved_vq}. Additionally, Fig. \ref{vq_cells} depicts the resulted Voronoi discretisation after implementing the VQ algorithm. Note that with VQ only those parts of the state space that are relevant to satisfying the property are accurately partitioned.  

Finally, we present the results of FVI method in Fig \ref{MDPs_solved_fvi} for the LTL formulas (\ref{omega}) and (\ref{ltl}). The FVI smoothing parameter is $h=0.18$ and the sampling time is $Z=25$ for both regions where both are empirically adjusted to have the minimum possible value for FVI to generate satisfying policies. The number of basis points also is set to be $100$, so the sample complexity\footnote{We do not sample the states in the product automaton that are associated to the accepting state of the automaton since when we reach the accepting state the property is satisfied and there is no need for further exploration. Hence, the last term is $ (|\mathcal{Q}|-1) $. However, if the property of interest produces an automaton that has multiple accepting states, then we need to sample those states as well.} of the FVI is equal to $100 \times Z \times |\mathcal{A}| \times (|\mathcal{Q}|-1)$. Note that in Table \ref{table:1}, in terms of timing, FVI outperforms the other methods. However, we have to remember that FVI is an approximate DP algorithm, which inherently needs an approximation of the transition probabilities. Therefore, as we have seen in Section \ref{FVIsection} in \eqref{montec}, for the set of basis points we need to sample the subsequent states. This reduces FVI applicability as it is often not possible in practice.

Additionally, both FVI and episodic VQ need careful hyper-parameter tuning to generate a satisfying policy, i.e., $h$ and $Z$ for FVI and $\Delta$ for VQ. The big merit of LCNFQ is that it does not need any external intervention. Further, as in Table \ref{table:1}, LCNFQ succeeds to efficiently generate a better policy compared to FVI and VQ. LCNFQ has less sample complexity while at the same time produces policies that are more reliable and also has better expected reward, i.e. higher probability of satisfying the given property. 

\section{Conclusion}
This paper proposes LCNFQ, the first RL algorithm to train Q-function in a continuous-state MDP such that the resulting traces satisfy a logical property. LCNFQ exploits the positive effects of generalization in neural networks while at the same time avoid the negative effects of disturbing previously learned experiences. This means that LCNFQ requires less experience and the learning process is highly data efficient which subsequently increases scalability and applicability of the proposed algorithm. LCNFQ is model-free, meaning that the learning only depends on the sample experiences that the agent gathered by interacting and exploring the MDP.  LCNFQ is \edit{successfully} tested in numerical examples to verify its performance and it outperformed the competitors.

\clearpage
\bibliographystyle{splncs}  
\bibliography{aamas_bib}  

\begin{thebibliography}{10}

\bibitem{doya}
Doya, K.:
\newblock Reinforcement learning in continuous time and space.
\newblock Neural computation \textbf{12}(1) (2000)  219--245

\bibitem{cdp}
Stachurski, J.:
\newblock Continuous state dynamic programming via nonexpansive approximation.
\newblock Computational Economics \textbf{31}(2) (2008)  141--160

\bibitem{cmac}
Sutton, R.S.:
\newblock Generalization in reinforcement learning: Successful examples using
  sparse coarse coding.
\newblock In: NIPS. (1996)  1038--1044

\bibitem{ormo}
Ormoneit, D., Sen, {\'S}.:
\newblock Kernel-based reinforcement learning.
\newblock Machine learning \textbf{49}(2) (2002)  161--178

\bibitem{ernst}
Ernst, D., Geurts, P., Wehenkel, L.:
\newblock Tree-based batch mode reinforcement learning.
\newblock JMLR \textbf{6}(Apr) (2005)  503--556

\bibitem{babuska}
Busoniu, L., Babuska, R., De~Schutter, B., Ernst, D.:
\newblock Reinforcement Learning and Dynamic Programming Using Function
  Approximators. Volume~39.
\newblock CRC press (2010)

\bibitem{apprx}
Hornik, K.:
\newblock Approximation capabilities of multilayer feedforward networks.
\newblock Neural networks \textbf{4}(2) (1991)  251--257

\bibitem{deepql}
Mnih, V., Kavukcuoglu, K., Silver, D., Rusu, A.A., Veness, J., Bellemare, M.G.,
  Graves, A., Riedmiller, M., Fidjeland, A.K., Ostrovski, G.,  et~al.:
\newblock Human-level control through deep reinforcement learning.
\newblock Nature \textbf{518}(7540) (2015)  529--533

\bibitem{tesauro}
Tesauro, G.:
\newblock {TD-Gammon}: A self-teaching {B}ackgammon program.
\newblock In: Applications of Neural Networks.
\newblock Springer (1995)  267--285

\bibitem{asynchronous}
Mnih, V., Badia, A.P., Mirza, M., Graves, A., Lillicrap, T., Harley, T.,
  Silver, D., Kavukcuoglu, K.:
\newblock Asynchronous methods for deep reinforcement learning.
\newblock In: ICML. (2016)  1928--1937

\bibitem{nfq}
Riedmiller, M.:
\newblock Neural fitted {Q} iteration-first experiences with a data efficient
  neural reinforcement learning method.
\newblock In: ECML. Volume 3720., Springer (2005)  317--328

\bibitem{cacla}
Van~Hasselt, H., Wiering, M.A.:
\newblock Reinforcement learning in continuous action spaces.
\newblock In: ADPRL, IEEE (2007)  272--279

\bibitem{sutton}
Sutton, R.S., Barto, A.G.:
\newblock Reinforcement learning: An introduction. Volume~1.
\newblock MIT press Cambridge (1998)

\bibitem{smith}
Smith, S.L., Tumov{\'a}, J., Belta, C., Rus, D.:
\newblock Optimal path planning for surveillance with temporal-logic
  constraints.
\newblock The International Journal of Robotics Research \textbf{30}(14) (2011)
   1695--1708

\bibitem{pol-sketch}
Andreas, J., Klein, D., Levine, S.:
\newblock Modular multitask reinforcement learning with policy sketches.
\newblock In: ICML. Volume~70. (2017)  166--175

\bibitem{AKLP10}
Abate, A., Katoen, J., Lygeros, J., Prandini, M.:
\newblock Approximate model checking of stochastic hybrid systems.
\newblock European Journal of Control \textbf{16}(6) (2010)  624--641

\bibitem{HSA17}
Haesaert, S., Soudjani, S., Abate, A.:
\newblock Verification of general markov decision processes by approximate
  similarity relations and policy refinement.
\newblock SIAM Journal on Control and Optimisation \textbf{55}(4) (2017)
  2333--2367

\bibitem{tkachev}
Tkachev, I., Mereacre, A., Katoen, J.P., Abate, A.:
\newblock Quantitative model-checking of controlled discrete-time {M}arkov
  processes.
\newblock Information and Computation \textbf{253} (2017)  1--35

\bibitem{wolf}
Wolff, E.M., Topcu, U., Murray, R.M.:
\newblock Robust control of uncertain {M}arkov decision processes with temporal
  logic specifications.
\newblock In: CDC, IEEE (2012)  3372--3379

\bibitem{topku}
Fu, J., Topcu, U.:
\newblock Probably approximately correct {MDP} learning and control with
  temporal logic constraints.
\newblock In: Robotics: Science and Systems X. (2014)

\bibitem{brazdil}
Br{\'a}zdil, T., Chatterjee, K., Chmel{\'\i}k, M., Forejt, V.,
  K{\v{r}}et{\'\i}nsk{\`y}, J., Kwiatkowska, M., Parker, D., Ujma, M.:
\newblock Verification of {M}arkov decision processes using learning
  algorithms.
\newblock In: ATVA, Springer (2014)  98--114

\bibitem{arxiv}
Hasanbeig, M., Abate, A., Kroening, D.:
\newblock Logically-constrained reinforcement learning.
\newblock arXiv preprint arXiv:1801.08099 (2018)

\bibitem{journal_arxiv}
Hasanbeig, M., Abate, A., Kroening, D.:
\newblock Certified reinforcement learning with logic guidance.
\newblock arXiv preprint arXiv:1902.00778 (2019)

\bibitem{hahn}
Hahn, E.M., Perez, M., Schewe, S., Somenzi, F., Trivedi, A., Wojtczak, D.:
\newblock Omega-regular objectives in model-free reinforcement learning.
\newblock arXiv preprint arXiv:1810.00950 (2018)

\bibitem{shield}
Alshiekh, M., Bloem, R., Ehlers, R., K{\"o}nighofer, B., Niekum, S., Topcu, U.:
\newblock Safe reinforcement learning via shielding.
\newblock arXiv preprint arXiv:1708.08611 (2017)

\bibitem{teacher}
Thomaz, A.L., Breazeal, C.:
\newblock Teachable robots: Understanding human teaching behavior to build more
  effective robot learners.
\newblock Artificial Intelligence \textbf{172}(6-7) (2008)  716--737

\bibitem{fulton}
Fulton, N., Platzer, A.:
\newblock Verifiably safe off-model reinforcement learning.
\newblock arXiv preprint arXiv:1902.05632 (2019)

\bibitem{fulton2}
Fulton, N.:
\newblock Verifiably Safe Autonomy for Cyber-Physical Systems.
\newblock PhD thesis, Carnegie Mellon University Pittsburgh, PA (2018)

\bibitem{fulton3}
Fulton, N., Platzer, A.:
\newblock Safe reinforcement learning via formal methods: Toward safe control
  through proof and learning.
\newblock In: Thirty-Second AAAI Conference on Artificial Intelligence. (2018)

\bibitem{ddl}
Platzer, A.:
\newblock Differential dynamic logic for hybrid systems.
\newblock Journal of Automated Reasoning \textbf{41}(2) (2008)  143--189

\bibitem{abbeel}
Moldovan, T.M., Abbeel, P.:
\newblock Safe exploration in {M}arkov decision processes.
\newblock arXiv preprint arXiv:1205.4810 (2012)

\bibitem{initial1}
Song, Y., Li, Y.b., Li, C.h., Zhang, G.f.:
\newblock An efficient initialization approach of {Q}-learning for mobile
  robots.
\newblock International Journal of Control, Automation and Systems
  \textbf{10}(1) (2012)  166--172

\bibitem{initial2}
Lope, J., Martin, J.:
\newblock Learning autonomous helicopter flight with evolutionary reinforcement
  learning.
\newblock In: International Conference on Computer Aided Systems Theory,
  Springer (2009)  75--82

\bibitem{stochastic}
Durrett, R.:
\newblock Essentials of stochastic processes. Volume~1.
\newblock Springer (1999)

\bibitem{puterman}
Puterman, M.L.:
\newblock {M}arkov decision processes: {D}iscrete stochastic dynamic
  programming.
\newblock John Wiley \& Sons (2014)

\bibitem{puterman2}
Cavazos-Cadena, R., Feinberg, E.A., Montes-De-Oca, R.:
\newblock A note on the existence of optimal policies in total reward dynamic
  programs with compact action sets.
\newblock Mathematics of Operations Research \textbf{25}(4) (2000)  657--666

\bibitem{watkins}
Watkins, C.J., Dayan, P.:
\newblock {Q}-learning.
\newblock Machine learning \textbf{8}(3-4) (1992)  279--292

\bibitem{multilayer}
Hornik, K., Stinchcombe, M., White, H.:
\newblock Multilayer feedforward networks are universal approximators.
\newblock Neural networks \textbf{2}(5) (1989)  359--366

\bibitem{nfq-1}
Riedmiller, M.:
\newblock Concepts and facilities of a neural reinforcement learning control
  architecture for technical process control.
\newblock Neural computing \& applications \textbf{8}(4) (1999)  323--338

\bibitem{nfq-2}
Lin, L.H.:
\newblock Self-improving reactive agents based on reinforcement learning,
  planning and teaching.
\newblock Machine learning \textbf{8}(3/4) (1992)  69--97

\bibitem{vectorq}
Gray, R.:
\newblock Vector quantization.
\newblock IEEE ASSP Magazine \textbf{1}(2) (1984)  4--29

\bibitem{voronoi}
Lee, I.S., Lau, H.Y.:
\newblock Adaptive state space partitioning for reinforcement learning.
\newblock Engineering applications of artificial intelligence \textbf{17}(6)
  (2004)  577--588

\bibitem{gordon}
Gordon, G.J.:
\newblock Stable function approximation in dynamic programming.
\newblock In: Machine Learning.
\newblock Elsevier (1995)  261--268

\bibitem{pnueli}
Pnueli, A.:
\newblock The temporal logic of programs.
\newblock In: Foundations of Computer Science, IEEE (1977)  46--57

\bibitem{sickert}
Sickert, S., Esparza, J., Jaax, S., K{\v{r}}et{\'\i}nsk{\`y}, J.:
\newblock Limit-deterministic {B{\"u}chi} automata for linear temporal logic.
\newblock In: CAV, Springer (2016)  312--332

\bibitem{sickert2}
Sickert, S., K{\v{r}}et{\'\i}nsk{\`y}, J.:
\newblock {MoChiBA}: Probabilistic {LTL} model checking using
  limit-deterministic {B{\"u}chi} automata.
\newblock In: ATVA, Springer (2016)  130--137

\bibitem{onehot}
Harris, D., Harris, S.:
\newblock Digital design and computer architecture.
\newblock Morgan Kaufmann (2010)

\bibitem{rprop}
Riedmiller, M., Braun, H.:
\newblock A direct adaptive method for faster backpropagation learning: The
  {RPROP} algorithm.
\newblock In: Neural networks, IEEE (1993)  586--591

\bibitem{hernandez}
Hern{\'a}ndez-Lerma, O., Lasserre, J.B.:
\newblock Further topics on discrete-time {M}arkov control processes.
  Volume~42.
\newblock Springer Science \& Business Media (2012)

\bibitem{montec}
Shonkwiler, R.W., Mendivil, F.:
\newblock Explorations in {M}onte {C}arlo Methods.
\newblock Springer Science \& Business Media (2009)

\bibitem{water_on_mars}
McEwen, A.S., Dundas, C.M., Mattson, S.S., Toigo, A.D., Ojha, L., Wray, J.J.,
  Chojnacki, M., Byrne, S., Murchie, S.L., Thomas, N.:
\newblock Recurring slope lineae in equatorial regions of {M}ars.
\newblock Nature Geoscience \textbf{7}(1) (2014) ~53

\end{thebibliography}

\end{document}